\documentclass[11pt]{article}
\usepackage[utf8]{inputenc}
\usepackage[margin=1in]{geometry}

\usepackage[round,authoryear]{natbib}
\usepackage{color}
\definecolor{waymogreen}{RGB}{0,232,157}
\definecolor{darkwaymogreen}{RGB}{0,190,129}
\definecolor{waymoblue}{RGB}{0,120,255}
\definecolor{natureblue}{RGB}{2,62,138}
\usepackage[colorlinks,citecolor=natureblue,linkcolor=waymoblue]{hyperref}
\usepackage{graphicx}
\usepackage{amsthm}
\usepackage{amsmath}
\usepackage{amssymb}
\usepackage{placeins}
\usepackage{textcomp,gensymb}
\usepackage{caption, subcaption}
\usepackage{comment}
\usepackage{booktabs}
\usepackage{adjustbox}
\usepackage{tikz}
\usepackage{url}
\usetikzlibrary{bayesnet}
\usepackage{multirow}
\usepackage{afterpage}
\usepackage{lscape}
\usepackage{lineno}
\usepackage{floatrow}
\usepackage{parskip} 
\usepackage{amsmath, mathrsfs}
\usepackage[font=small]{caption}
\usepackage{mwe}
\usepackage[hang,flushmargin]{footmisc}
\usepackage{float}
\floatstyle{plaintop}
\restylefloat{table}
\usepackage{pifont}% http://ctan.org/pkg/pifont
\newcommand{\cmark}{\ding{51}}
\newcommand{\xmark}{\ding{55}}
\makeatletter
\newcommand*\bigcdot{\mathpalette\bigcdot@{.5}}
\newcommand*\bigcdot@[2]{\mathbin{\vcenter{\hbox{\scalebox{#2}{$\m@th#1\bullet$}}}}}

\patchcmd{\NAT@test}{\else \NAT@nm}{\else \NAT@hyper@{\NAT@nm}}{}{}
\makeatother

\usepackage{array}
\newcolumntype{L}[1]{>{\raggedright\let\newline\\\arraybackslash\hspace{0pt}}m{#1}}
\newcolumntype{C}[1]{>{\centering\let\newline\\\arraybackslash\hspace{0pt}}m{#1}}
\newcolumntype{R}[1]{>{\raggedleft\let\newline\\\arraybackslash\hspace{0pt}}m{#1}}

\usepackage{setspace}
\usepackage{titlesec}

\titleformat*{\section}{\large\bfseries}
\titleformat*{\subsection}{\bfseries}
\titleformat*{\subsubsection}{\bfseries}
\titleformat*{\paragraph}{\itshape}
\titleformat*{\subparagraph}{\bfseries}

\title{Measuring Surprise in the Wild}
\author{Azadeh Dinparastdjadid\\
\texttt{azadehd@waymo.com}
\and Isaac Supeene\\
\texttt{isupeene@waymo.com}
\and Johan Engstr\"{o}m\\
\texttt{jengstrom@waymo.com}}

\date{}

\begin{document}
%\singlespacing
\onehalfspacing
\maketitle
\thispagestyle{empty}

\begin{abstract}
\noindent
The quantitative measurement of how and when we experience surprise has mostly remained limited to laboratory studies, and its extension to naturalistic settings has been challenging. Here we demonstrate, for the first time, how computational models of surprise rooted in cognitive science and neuroscience combined with state-of-the-art machine learned generative models can be used to detect surprising human behavior in complex, dynamic environments like road traffic. In traffic safety, such models can support the identification of traffic conflicts, modeling of road user response time, and driving behavior evaluation for both human and autonomous drivers. We also present novel approaches to quantify surprise and use naturalistic driving scenarios to demonstrate a number of advantages over existing surprise measures from the literature. Modeling surprising behavior using learned generative models is a novel concept that can be generalized beyond traffic safety to any dynamic real-world environment. 
\end{abstract}

\newpage
\setcounter{page}{1}
%\linenumbers
\nolinenumbers

% Main text
\section{Introduction}\label{intro}

\noindent

Amazed, astonished, astounded, and flabbergasted! We have all experienced these terms through surprising experiences in our lives such as entering a surprise birthday party, or jumping when a balloon from said birthday party unexpectedly popped in the middle of the night. Our experiences with surprise carry positive or negative emotions at varying levels of intensity. Thanks to surprise, we are enthralled by the plot twists of a good story (\cite{aristotle_2013, perez2020jon}), mesmerized by a close game of sports (\cite{antony2021behavioral}), and captivated by an emotional piece of music (\cite{cheung2019uncertainty, gold2019predictability, shany2019surprise}). But what does it really mean to be surprised? While the concept seems obvious, it has prompted research dating as far back as Aristotle describing surprise as a mental and behavioral phenomenon (about 350 B.C.; see \cite{aristotle_1980}). For example, with music, both surprise and uncertainty resolution have been shown to correlate with the emotional experience and pleasantness of music (\cite{leonard1956emotion, huron2008sweet, shany2019surprise, cheung2019uncertainty}). Surprise is also a key aspect of humor. Stand up comedians often start their jokes with the \emph{set-up}, creating a certain expectation, and then deliver the \emph{punchline} which violates the initial expectation. This shifting and dissipating of our expectations is said to make jokes amusing (\cite{morreall2012philosophy, raccah2016humour, de2020jokes}).

The key role of expectations and expectation violations (i.e., surprise) in the context of road traffic has long been acknowledged (\cite{alexander1986driver, martens2007failure,theeuwes1996visual, theeuwes1995self, rasanen1998attention}). \cite{glauz1980application} explicitly called out the notion of atypical / unusual road user actions in their definition of traffic conflicts: “a traffic conflict is a traffic event involving two or more road users, in which one user performs some atypical or unusual action, such as a change in direction or speed, that places another user in jeopardy of a collision unless an evasive maneuver is undertaken. (p. 5)”  In line with this, \cite{tageldin2016developing} showed that traffic conflict indicators based on sudden evasive action were better at identifying pedestrian conflicts and estimating their severity than traditional proximity indicators like time to collision. \cite{bagdadi2011jerky} further demonstrated that jerky, abrupt road user behavior is indicative of increased crash risk. The formal ISO definitions of a crash and a near crash (\cite{ISO_2018}) require that a true crash or near-crash be “not premeditated", and the colloquial term “accident" mirrors this emphasis on unexpectedness, and hence surprise. Predictability has also been proposed as a key principle of good autonomous vehicle (AV) driving behavior (\cite{de2021driverless}). Despite the important conceptual role surprise plays in traffic safety research, there is no precise quantitative definition or computational model of surprise in this domain. This paper is, to our knowledge, the first attempt to show how computational models of surprise rooted in cognitive science and neuroscience can be generalized and used to detect surprising human behavior in complex, dynamic environments like road traffic.

So how can surprise be operationalized? The quantitative study and modeling of surprise has attracted researchers across many scientific disciplines including psychology (e.g., \cite{mellers1997decision, reisenzein2000exploring}), neuroscience (e.g., \cite{preuschoff2011pupil}), and artificial intelligence (e.g., \cite{macedo2009artificial, berseth2019smirl}. Surprise plays a key role in models of learning and memory (\cite{sutton1998reinforcement, sinclair2018surprise}), exploration (\cite{schwartenbeck2013exploration}), visual attention (\cite{itti2009bayesian}), and demarcating events in the continuous flow of time (\cite{franklin2020structured}). Such research has been published under headings such as Bayesian inference, active inference, the free energy principle, belief-updating, prediction error, schema revision, and many others (e.g., \cite{parr2022active, reisenzein2019cognitive}).

\cite{itti2009bayesian}, proposed two essential components for any principled definition of surprise: 1) the presence of uncertainty, and 2) subjectivity. Uncertainty depends on factors such as missing information, limited computing resources, or intrinsic stochasticity leading to a non-deterministic world for a given observer. On the other hand, surprise is always tied to the expectations of a specific observer and the same observation may cause different amounts of surprise for different observers. Moreover, the same observer may experience different amounts of surprise at different times (\cite{itti2009bayesian}). These two ingredients point towards a probabilistic setting in which surprise can be generally conceptualized as a violation of an agent’s subjective belief about the state of the world, where a belief is operationalized as a probability distribution over states (\cite{kaelbling1998planning}).

Given that surprise is subjective and experienced from the perspective of a particular agent, the notion of a \emph{generative model} becomes a core concept in operationalizing surprise. In simple terms, a generative model is the brain’s internal representation of the world that generates an agent's expectations of sensory signals (\cite{friston2001dynamic, bruineberg2018free}). Computational models related to decision-making, learning, perception, and memory typically assume that humans implicitly perceive their sensory observations as probabilistic outcomes of a generative model with hidden variables (\cite{findling2021imprecise, fiser2010statistically, friston2010free, gershman2017computational, liakoni2021learning, soltani2019adaptive, angela2005uncertainty}). The actual dynamics of the world may be different from those inferred by the agent based on its generative model (\cite{modirshanechi2022taxonomy}). This leads to the definition of the \emph{generative process} which represents the true causal structure of the world that generates the sensory information that agents observe. The generative model can be seen as an  approximation of the generative process which may not always be accurate (\cite{bruineberg2018free}).

While there seems to be general agreement in the literature on the conceptualization of surprise, and that it is experienced in relation to subjective, probabilistic beliefs, there are many different proposals on how to operationalize surprise. For present purposes, following \cite{modirshanechi2022taxonomy}, we distinguish between three general types of surprise measures: (1) \emph{probabilistic mismatch surprise}, (2) \emph{belief mismatch surprise} and (3) \emph{observation-mismatch surprise}. Probabilistic mismatch surprise compares an observed state to a prior belief. In this setting, an observation that had a low probability under the observer's prior belief will lead to an experience of surprise. One existing computational surprise model in this category is Shannon surprise, also known as surprisal (\cite{shannon1948mathematical}). As described in Equation~\ref{eq:surprisal} below, surprisal is defined as the negative log probability of an event under some prior probability distribution $P$. Thus if an event $x$ has a low probability under $P$, surprisal will be high.

\begin{equation}
    S(x; P) =  -\log(P(x))
    \label{eq:surprisal}
\end{equation}

Other examples of probabilistic mismatch surprise measures include Bayes factor surprise (\cite{liakoni2021learning}), and state prediction error (\cite{glascher2010states}).

The second category, belief mismatch surprise, compares two belief distributions. An example of this category is Bayesian surprise (\cite{itti2009bayesian}). As described in \cite{itti2009bayesian}, the prior probability distribution $\{P(M)_{M\in\mathcal{M}}\}$ is defined over the hypotheses or models $M$ in a model space $\mathcal{M}$. The likelihood function $P(D|M)$ is associated with each of the hypotheses or models $M$ and it quantifies the likelihood of any data observation $D$, assuming that a particular model $M$ is correct (\cite{itti2009bayesian}). According to Bayes theorem,

\begin{equation}
    \forall M \in \mathcal{M}, P(M|D) = \frac{P(D|M)P(M)}{P(D)}
\end{equation}

the prior distribution of beliefs $\{P(M)_{M\in\mathcal{M}}\}$ will change to the posterior distribution $\{P(M|D)_{M\in\mathcal{M}}\}$ with the observation of new data $D$. The prior and posterior belief distributions reflect subjective probabilities across the possible outcomes (\cite{kaelbling1998planning}) and Bayesian surprise is the difference between the posterior and prior distribution, which in \cite{itti2009bayesian} is quantified using the Kullback-Leibler (KL) divergence. Other examples of belief mismatch surprise measures include postdictive surprise (\cite{kolossa2015computational}), confidence corrected surprise (\cite{faraji2018balancing}), and free energy (\cite{friston2010free, friston2017active, gershman2019does}). 

\cite{modirshanechi2022taxonomy} also proposed a third category called observation-mismatch surprise, which generally refers to a mismatch between a predicted and an actual observation. Examples of this category are absolute and squared error surprise (\cite{prat2021human}), and the unsigned reward prediction error (\cite{hayden2011surprise, pearce1980model, rouhani2021signed, talmi2013feedback}). 

These existing  computational surprise measures have typically been applied in laboratory experiments and their extension to naturalistic settings (\cite{antony2021behavioral}) has been challenging,  particularly in complex domains like traffic safety. Existing work towards applying surprise measures in the real world include
\citeauthor{engstrom2018great}'s \citeyearpar{engstrom2018great}, general framework for understanding driving based on surprise minimization, and \citeauthor{bianchi2020drivers}'s \citeyearpar{bianchi2020drivers}, computational model of expectation mismatches developed to predict human driver responses to silent automation (adaptive cruise control) failures (see also \cite{victor2018automation}). In a similar vein, \cite{engstrom2022modeling} proposed a framework and a specific model for road user response timing based on surprise and Bayesian belief updating. 

An important prerequisite for real world surprise measures is generative models that can be applied to naturalistic settings. Generative models are typically defined analytically, for example by a Partially Observable Markov Decision Process (POMDP) for discrete time problems, or stochastic differential equations for continuous time problems (\cite{parr2022active}; Chapter 4). To scale to complex real-world problems like road traffic, machine learned function approximators like neural networks can be used as generative models (\cite{tschantz2020scaling}).

In this paper we describe a novel approach for quantifying surprising road user behavior based on behavior predictions obtained from a machine-learned generative model. The main contributions of this paper are (i) novel ways to quantify surprise using state-of-the-art machine learned generative models, and (ii) demonstrating for the first time, to the best of our knowledge, how surprising human behavior can be objectively detected in complex, dynamic environments like road traffic. We demonstrate the application of our novel surprise measures along with two existing measures of surprise (surprisal and Bayesian surprise) using naturalistic driving examples, and discuss how our surprise measures can be used for several different road traffic applications, including the identification of traffic conflicts, the modeling of road user response time, and driving behavior evaluation for both human and autonomous drivers.

\section{Results}

In our operationalization of road user surprise, beliefs are represented as the output of a generative model. Our generative model is an evolution of the \emph{Multipath} model (\cite{chai2019multipath}) using the \emph{Wayformer} encoder (\cite{nayakanti2022wayformer}), which produces probabilistic predictions about how a traffic situation will play out. These predictions are based on an understanding of the static and dynamic world context including road semantics (e.g., lane connectivity, stop lines), traffic light information, and past observations of other agents. The model’s outputs include (1) a set of discrete trajectories that are both weighted and parsimonious, covering the space of likely outcomes, and (2) the likelihood of any trajectory (\cite{chai2019multipath}). In simple terms, the model learns to predict probability distributions over future road user position by observing real-world traffic.

The model space of these beliefs can vary based on the considered level of abstraction. Higher levels of abstraction will include hypotheses about the possible action space (e.g., pass, yield, decelerate, accelerate) creating a discrete probability distribution for the belief. Lower levels of abstraction can include predictions over continuous variables such as the lateral position of another road user, or the ego’s own position relative to the road edge at different time steps into the future. In our use case, the generative model’s predictions fall on the lower end of the spectrum providing belief distributions on lateral and longitudinal positions. 

\begin{figure}[t]
	\centering
		\includegraphics[width=1\textwidth]{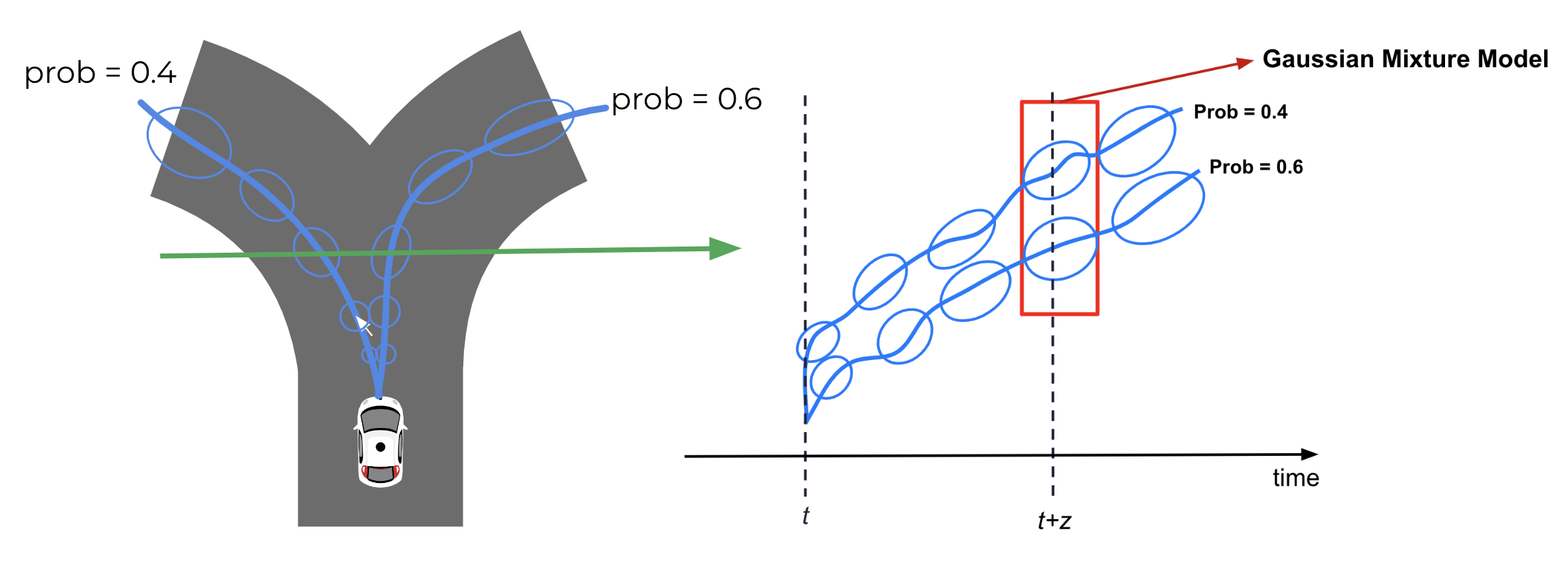}
	  \caption{Uncertainty in predictions at lower levels of abstraction. The generative model produces a time series of predictions, at time $t$.}
	  \label{fig:GMM_fig}
\end{figure}

Figure~\ref{fig:GMM_fig} provides a simplified illustration of the output of the generative model which is a 2-d continuous distribution over future positions. The model represents two types of uncertainty: 1) uncertainty about the agent’s intended route, and 2) uncertainty about the state of the agent at each timestamp on a given trajectory. The probability of taking the different paths in Figure~\ref{fig:GMM_fig} reflects the first type of uncertainty. The blue ovals illustrate the 2-d Gaussian distributions of the agent’s future position at each timestamp, and demonstrate the uncertainty at each timestamp.

As we predict further into the future, uncertainty about the state of the agent will increase, as shown by the increasing size of the blue ovals. Considering the current time $t$, if we look at the predictions about the future state of an agent along a certain trajectory 5s into the future, we would expect more uncertainty than when looking at predictions 0.5s into the future. Combining the Gaussians from the different trajectories at a particular timestamp will produce a Gaussian Mixture Model (\emph{GMM}) as indicated by the red box in Figure~\ref{fig:GMM_fig}. These GMMs were used as the belief distributions for measuring surprise. 

\begin{figure}
  \ffigbox[\textwidth]{
    \begin{subfloatrow}[1]\useFCwidth
      \fcapside[\FBwidth]{\includegraphics[width=0.7\textwidth]{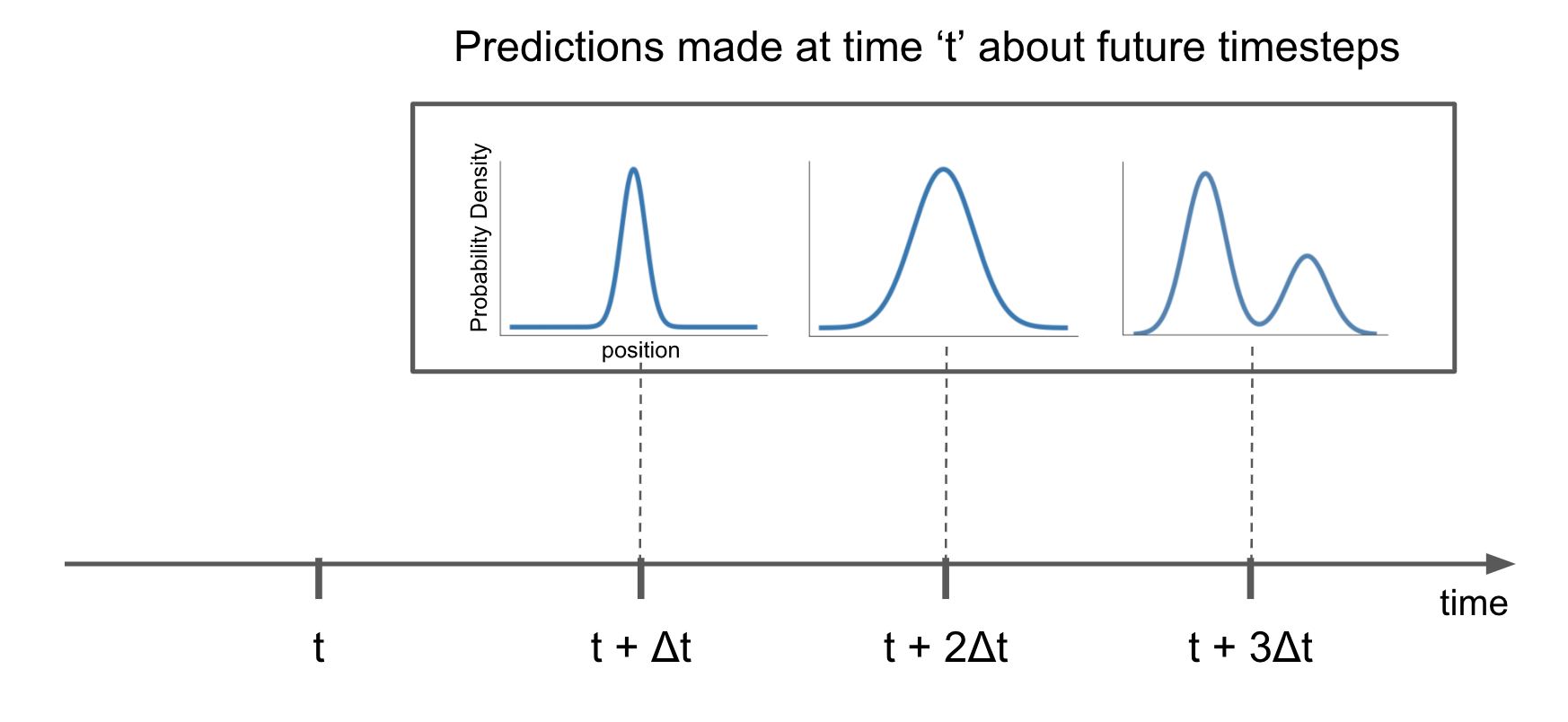}}{\caption{}\label{fig:concept_illustration_a}}
    \end{subfloatrow}
    \begin{subfloatrow}[1]\useFCwidth
      \fcapside[\FBwidth]{\includegraphics[width=0.65\textwidth]{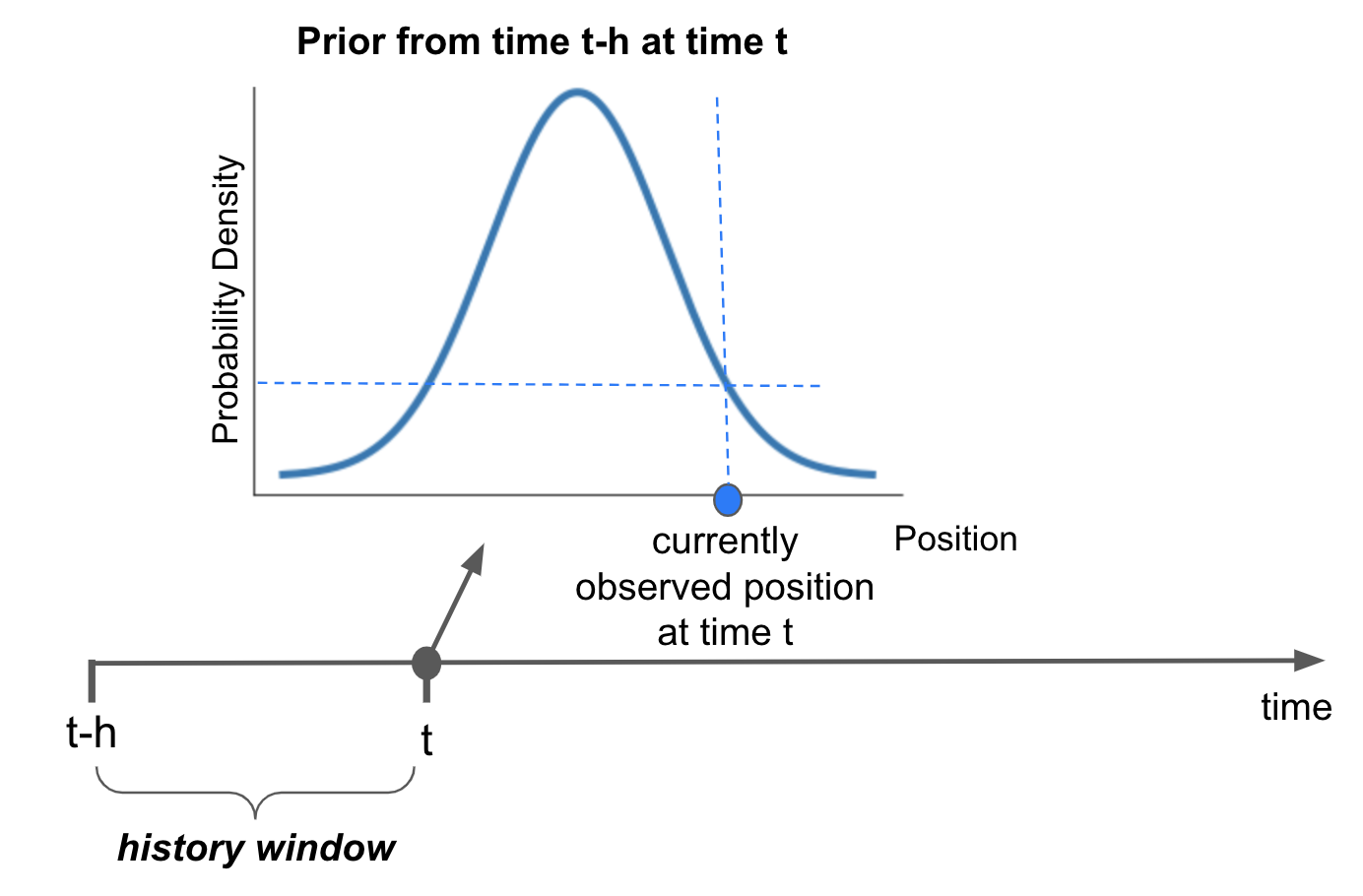}}{\caption{}\label{fig:concept_illustration_b}}
    \end{subfloatrow}
    \begin{subfloatrow}[1]\useFCwidth
      \fcapside[\FBwidth]{\includegraphics[width=0.7\textwidth]{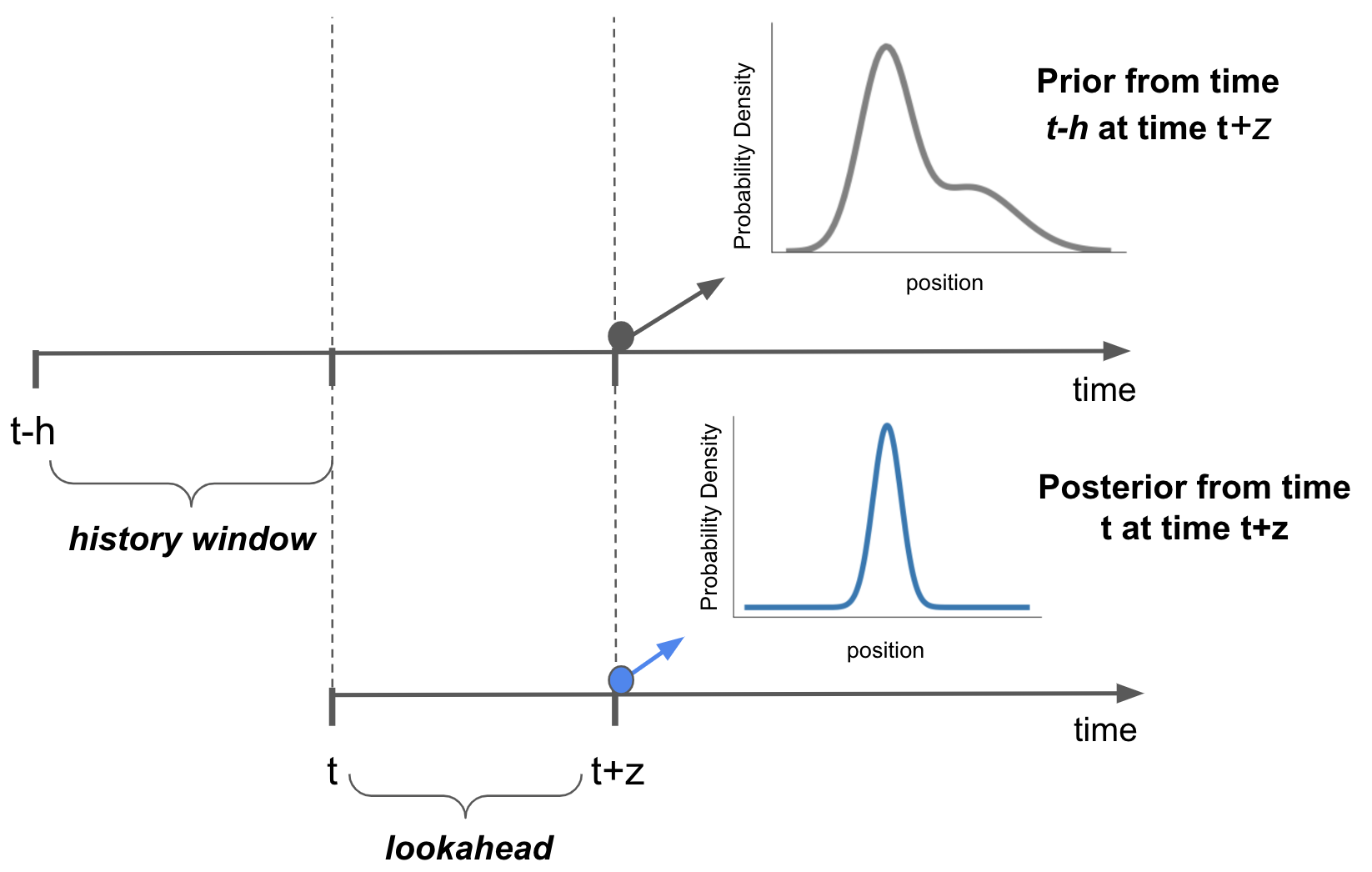}}{\caption{}\label{fig:concept_illustration_c}}
    \end{subfloatrow}
  }
  {\caption{Schematic illustration of probabilistic mismatch and belief mismatch surprise. a) the introduction of a time series of predictions from time $t$. On the miniature distributions, the y-axis is a probability density, and the x-axis is the road user's position. b) visualization of probabilistic mismatch surprise and the introduction of history window, $h$. c) visualization of belief mismatch surprise and the introduction of the lookahead time, $z$.
}
\label{fig:concept_illustration}}

\end{figure}

When discussing the belief distributions we should distinguish between the time the prediction was made and the time the prediction is about. Considering a timestep of ${\Delta}t$, Figure~\ref{fig:concept_illustration_a} illustrates a time series of predictions made at time $t$ about future timestamps $t+{\Delta}t$, $t+2{\Delta}t$, $t+3{\Delta}t$. The belief distribution at each of these timestamps (e.g., at $t+{\Delta}t$) is a Gaussian Mixture Model that was generated at time $t$.

The prior belief distribution is the common denominator between probabilistic mismatch and belief mismatch surprise and it requires the introduction of a new parameter, the \emph{history window}, $h$. As illustrated in Figure~\ref{fig:concept_illustration_b}, the history window represents how far back in time the prior belief was generated. In probabilistic mismatch surprise, an observation at time $t$ is compared to the prior belief distribution made at time $t-h$ about time $t$. Although the generative model output at time $t-h$ produces a time series of predictions at future timestamps, we, in this case, only consider the predictions for time $t$.

In the context of belief mismatch surprise, we must consider both prior and posterior belief distributions. This creates the need to introduce a second parameter, the \emph{lookahead time}, $z$, which represents how far into the future we predict. In belief mismatch surprise, predictions are generated at two different points in time: time $t - h$ for the generation of the prior, and time $t$ for the posterior. While the prior and posterior come from different timestamps, they are both about the same point in time, $t + z$, as illustrated in Figure~\ref{fig:concept_illustration_c}. We then compare the belief distributions about the road user's future position at time $t + z$ to measure surprise.

Existing measures of surprise such as surprisal, and information theory in general, have either implicitly or explicitly assumed discrete probability distributions (\cite{marsh2013introduction}). For road traffic, this would translate to, for example, a generative model that outputs pass/yield predictions while many applications such as traffic safety and driving may include operating over continuous probability distributions. To address this issue, we have proposed a series of new surprise metrics to accommodate continuous belief distributions coming from generative models like ours. In the Methods section, we describe these newly proposed surprise measures in more detail, and discuss the benefits of each relative to existing surprise measures such as Bayesian surprise and surprisal. In the next section, we discuss the application of four surprise measures, including our own novel measures, to real world driving events.

\subsection{Example applications to real-world driving scenarios}

\begin{figure*}
        \centering
        \begin{subfigure}[b]{0.475\textwidth}
            \centering
            \includegraphics[width=2.5in]{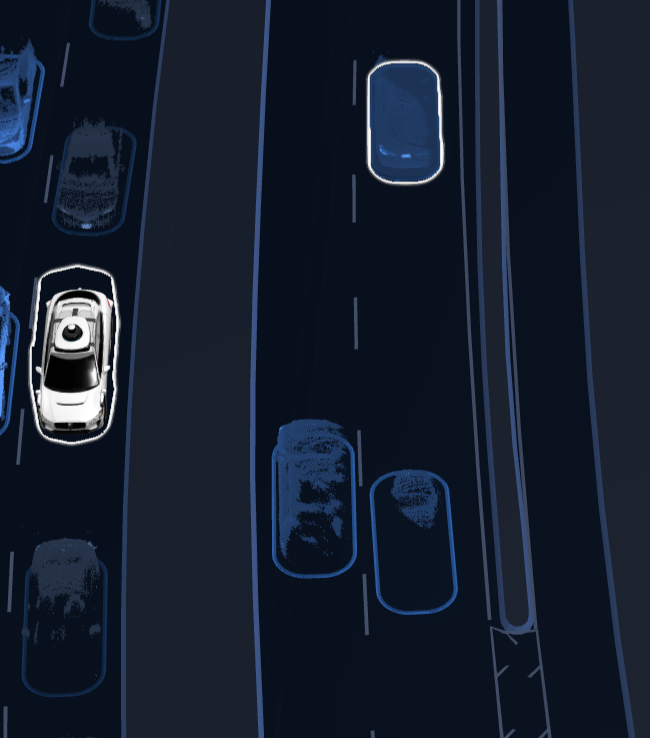}
            \caption{}   
            \label{fig:scenarios_pics_a}
        \end{subfigure}
        \hfill
        \begin{subfigure}[b]{0.475\textwidth}  
            \centering 
            \includegraphics[width=2.5in]{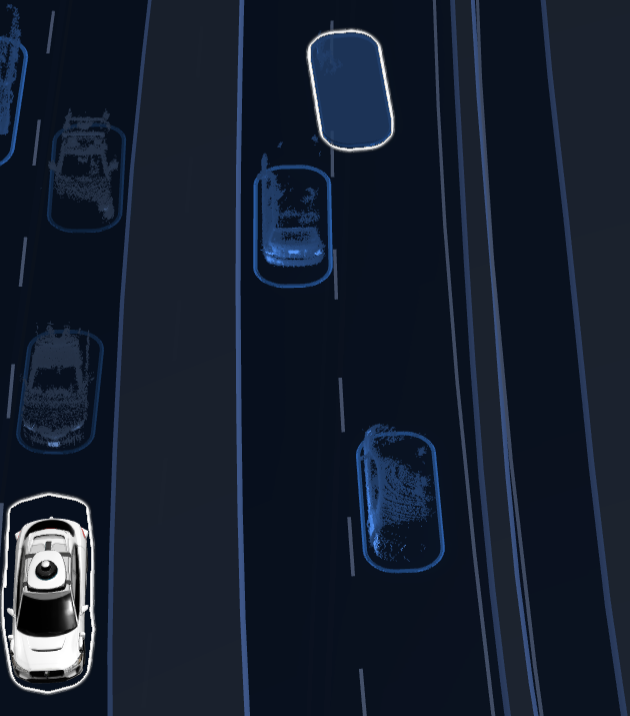}
            \caption{}  
            \label{fig:scenarios_pics_b}
        \end{subfigure}
        \vskip\baselineskip
        \begin{subfigure}[b]{0.475\textwidth}   
            \centering 
            \includegraphics[width=2.5in]{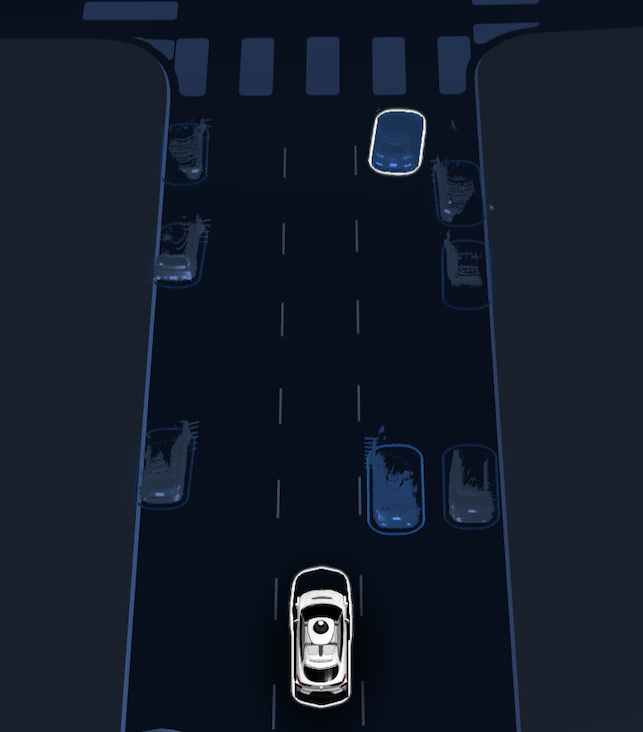}
            \caption{}
            \label{fig:scenarios_pics_c}
        \end{subfigure}
        \hfill
        \begin{subfigure}[b]{0.475\textwidth}   
            \centering 
            \includegraphics[width=2.48in]{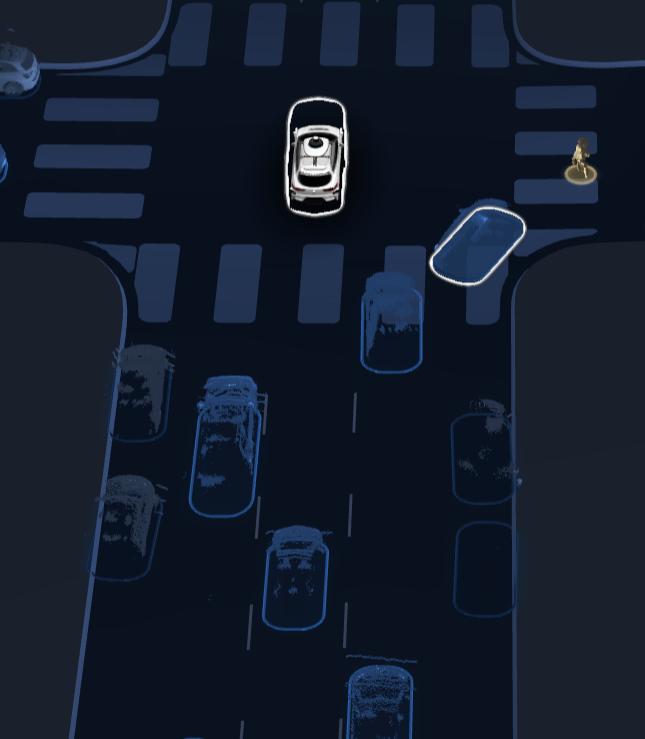}
            \caption{}
            \label{fig:scenarios_pics_d}
        \end{subfigure}
        \caption{Example of a laterally surprising behavior. a) normal driving prior to cut-in, b) laterally surprising cut-in event initiated by highlighted vehicle on the right, c) normal driving prior to surprising braking, d) longitudinally surprising stopping event initiated by highlighted vehicle turning right. The white vehicle is the Waymo.}
        \label{fig:scenarios_pics}
    \end{figure*}

For this proof of concept demonstration, we used naturalistic driving data collected by Waymo vehicles, which are AVs equipped with a wide range of sensors for perceiving the external driving environment. While driving on the road, Waymo vehicles can record interactions between other vehicles in their vicinity using their sensors (for example, the perception data for the scenario in Figure~\ref{fig:scenarios_pics_a} originates from the white Waymo vehicle). Using this data, we chose two events in which a laterally or longitudinally surprising maneuver was initiated by one of the other vehicles, the \emph{initiator}. Due to the surprising behavior of the initiator, another vehicle, the \emph{responder}, will need to perform an evasive maneuver (e.g., hard brake, swerve) to avoid a collision. Laterally surprising events such as surprising cut-ins and aborted lane changes involve an unexpected and abrupt lateral movement from the initiator. In longitudinally surprising events, the initiator performs an unexpected longitudinal maneuver such as sudden hard brakes, or unexpected accelerations/decelerations. To measure surprise in each of these categories, our surprise metric, which is based on position, was decomposed into its lateral and longitudinal components by transforming the coordinates to a body-frame reference. The result is a lateral or longitudinal time series surprise signal with peaks referring to surprising lateral or longitudinal behavior. 

Four surprise measures were applied to these two events. Two of these surprise measures, surprisal and Bayesian surprise, are based on existing literature but their application to the road traffic domain is novel. We also applied two new surprise measures to these events: Residual Information which belongs to the probabilistic mismatch category, and Antithesis which is a  belief mismatch surprise measure.

The first example, in Figures~\ref{fig:scenarios_pics_a} and \ref{fig:scenarios_pics_b}, is a laterally surprising event involving the highlighted vehicle suddenly cutting in front of the vehicle on its left. The second example, in Figures~\ref{fig:scenarios_pics_c} and \ref{fig:scenarios_pics_d}, is a longitudinally surprising event where the lead vehicle (highlighted vehicle), abruptly stopped and braked during a right turn, requiring a response from the following vehicle.

\begin{figure*}
        \centering
        \begin{subfigure}[b]{0.475\textwidth}
            \centering
            \includegraphics[width=\textwidth]{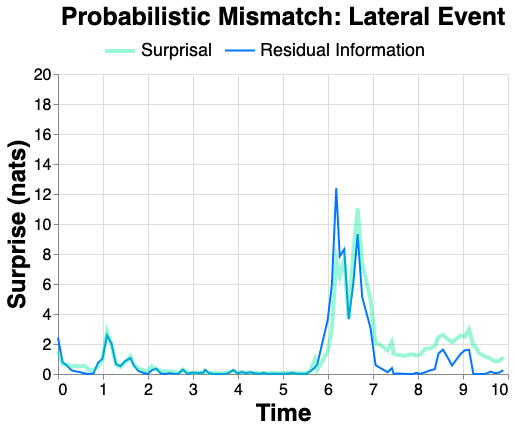}
            \caption{}
            \label{fig:results_surprise_a}
        \end{subfigure}
        \hfill
        \begin{subfigure}[b]{0.475\textwidth}  
            \centering 
            \includegraphics[width=\textwidth]{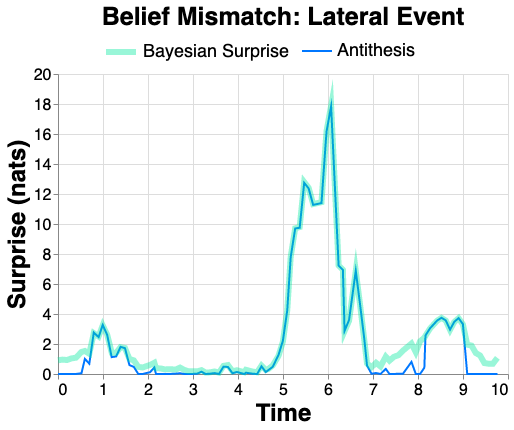}
            \caption{}
            \label{fig:results_surprise_b}
        \end{subfigure}
        \vskip\baselineskip
        \begin{subfigure}[b]{0.475\textwidth}   
            \centering 
            \includegraphics[width=\textwidth]{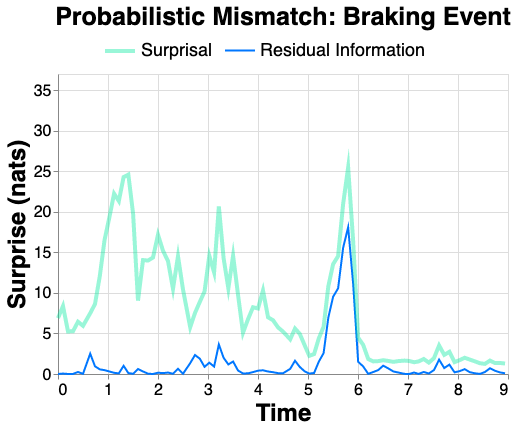}
            \caption{}
            \label{fig:results_surprise_c}
        \end{subfigure}
        \hfill
        \begin{subfigure}[b]{0.475\textwidth}   
            \centering 
            \includegraphics[width=\textwidth]{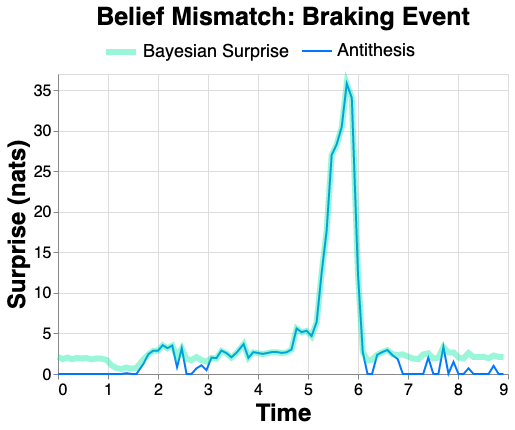}
            \caption{}
            \label{fig:results_surprise_d}
        \end{subfigure}
        \caption{Application of 4 surprise measures to a laterally surprising cut-in example (a), and (b), and a longitudinally surprising braking event (c), and (d). a, and c) probabilistic mismatch surprise measures including surprisal and Residual Information, b, and d) belief mismatch surprise measures including Bayesian surprise and Antithesis. For Antithesis and Bayesian surprise we used, $h=2$s, and $z=0.2$s, and $h=1$s for surprisal and Residual Information.}
        \label{fig:results_surprise}
    \end{figure*}

Figure~\ref{fig:results_surprise} demonstrates the application of the four surprise measures to the two example events. The expectation was to observe a visible peak in the surprise time series signal across these different surprise measures around 5s for the lateral cut-in and 5.5s for the surprising hard brake. The history window ($h$) and lookahead ($z$) parameters can be adjusted based on application needs. For demonstration purposes, we used $h=2$s, and $z=0.2$s for Antithesis and Bayesian surprise and $h=1$s for surprisal and Residual Information.

Figures~\ref{fig:results_surprise_a} and \ref{fig:results_surprise_c} compare the two probabilistic mismatch measures, surprisal and Residual Information. As discussed in the Methods section, the zero-floor issue with surprisal is evident, especially with the longitudinally surprising braking event in Figure~\ref{fig:results_surprise_c}. Figures~\ref{fig:results_surprise_b} and \ref{fig:results_surprise_d} compare the two belief mismatch measures, Bayesian surprise and Antithesis. As detailed in the Methods section, Antithesis measures the increased likelihood of a previously unexpected outcome such as a surprising cut-in or a sudden braking event, while silencing unsurprising information. Based on this, our expectation was for Antithesis to be zero more often than Bayesian surprise, which is supported by Figures~\ref{fig:results_surprise_b} and \ref{fig:results_surprise_d}.

\subsection{Effect of Parameters}

\begin{figure*}
        \centering
        \begin{subfigure}[b]{1.0\textwidth}
            \centering
            \includegraphics[width=\textwidth]{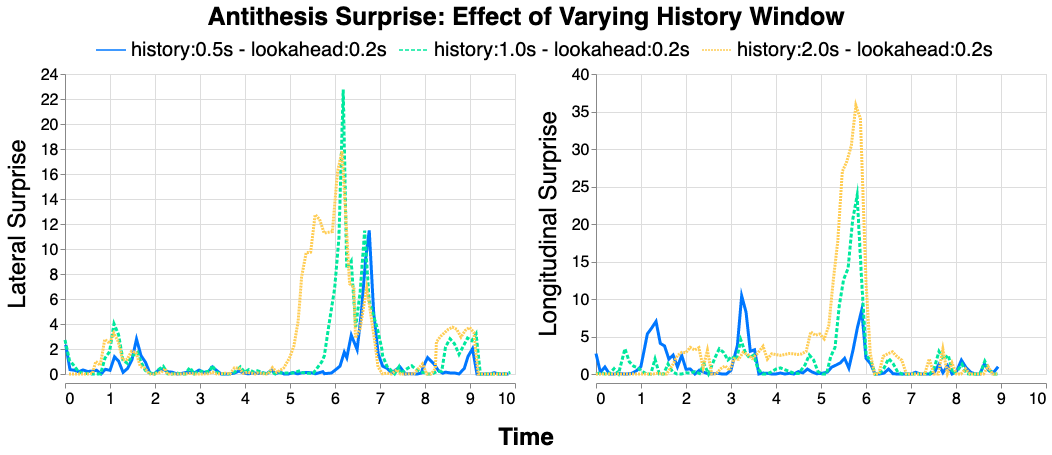}
            \subfloat[\label{fig:parameter_effect_a}]{\hspace{.45\linewidth}}
            \subfloat[\label{fig:parameter_effect_b}]{\hspace{.45\linewidth}}
        \end{subfigure}
        \vskip\baselineskip
        \begin{subfigure}[b]{1.0\textwidth}   
            \centering 
            \includegraphics[width=\textwidth]{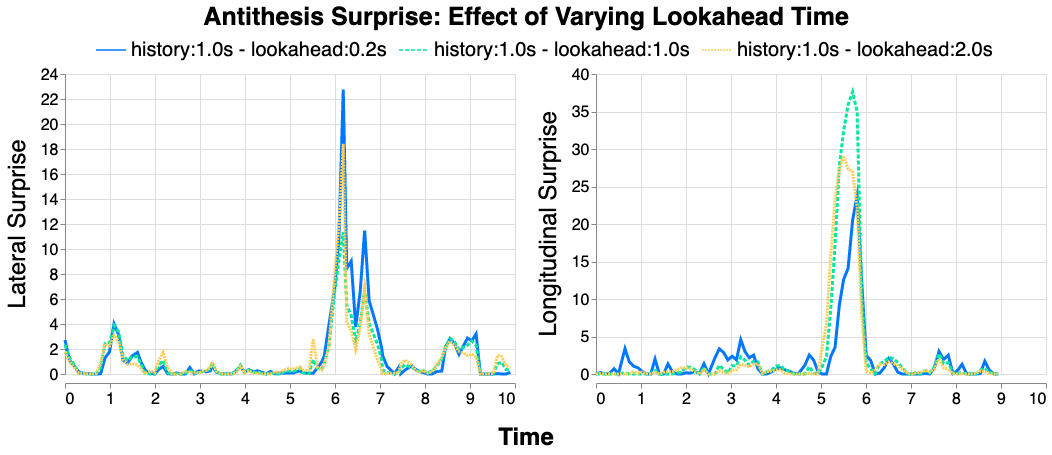}
            \subfloat[\label{fig:parameter_effect_c}]{\hspace{.45\linewidth}}
            \subfloat[\label{fig:parameter_effect_d}]{\hspace{.45\linewidth}}
        \end{subfigure}
        \caption{The effect of varying history window and lookahead on antithesis surprise. a and b) varying history window with a constant lookahead for laterally and longitudinally surprising events. c and d) varying lookahead with a constant history window  for laterally and longitudinally surprising events.}
        \label{fig:parameter_effect}
    \end{figure*}

Focusing on our novel Antithesis surprise measure, Figure~\ref{fig:parameter_effect} illustrates the effect of varying the parameters $h$ and $z$ on the surprise signal. In Figure~\ref{fig:parameter_effect_a} and ~\ref{fig:parameter_effect_b}, we kept the lookahead time ($z$) constant while varying the history window ($h$). It can be seen that in both examples, the smaller history windows resulted in lower magnitude peaks. This can be explained by the increased similarity between the prior and posterior as the time gap between the prior and posterior decreases. In addition, the surprise peaks for the larger history windows started earlier. In Figure~\ref{fig:parameter_effect_c}, when keeping the history window constant, the smallest lookahead time (0.2s) led to the highest magnitude peak. However, in Figure~\ref{fig:parameter_effect_d}, this lookahead value had the smallest peak, and the intermediate lookahead value (1s) had the largest peak. 

Based on these two events, there is no evident pattern to the effect of $z$. We believe a variety of factors determine the point in the future trajectory about which the new evidence is most informative; for example, the vehicle dynamics and action currently being undertaken affect our prior beliefs about an agent's trajectory at various timescales, and the nature of the surprising behavior may have either short- or long-term implications for the agent's future trajectory. These parameters can be tuned to accommodate particular applications and use cases. For example, when applying Antithesis to large trucks with more sluggish vehicle dynamics, increasing the lookahead window can be useful to amplify the surprise signal.

\section{Discussion}
Surprise is a pervasive phenomenon that plays a key role across a wide range of human behavior. Some contemporary models in cognitive science, neuroscience and machine learning even suggest surprise minimization as the single fundamental principle underlying behavior and cognition
(\cite{friston2010free, seth2016active, parr2022active}). Thus, quantifying surprise in real-world dynamic scenarios can advance our understanding of human behavior. In this paper, we demonstrated how surprising behavior can be measured in the complex, dynamic domain of road traffic. We used a machine-learned generative model to generate road user belief distributions which enabled both probabilistic mismatch and belief mismatch surprise measures. These included our two novel surprise measures, Residual Information and Antithesis, along with existing measures, surprisal (\cite{shannon1948mathematical}) and Bayesian surprise (\cite{itti2009bayesian}), which we applied to real-world driving examples. While the focus of this paper was on road traffic, our framework is generalizable to any domain where a generative model can be trained to generate predictions.

Our methods are applicable to any distribution, whether discrete or continuous, and explicitly consider the process of information acquisition over time. Moreover, while the generative model used in this paper made predictions at lower levels of abstraction (e.g., position), our methods can be generalized to more abstract states (e.g., pass/yield).

The precision, or inverse uncertainty, of the belief is a key aspect in surprise computation that we so far have not addressed explicitly in this paper. High precision (low uncertainty) corresponds to few potential outcomes and thus high confidence in a particular belief (or a limited set of beliefs). As a result, if an observation deviates from the prior belief, the potential for surprise is high if the prior belief had a high precision (e.g., a driver strongly believed that a pedestrian would yield at the crosswalk when the pedestrian suddenly crossed). Conversely, a distribution with low precision corresponds to many different potential outcomes and low confidence in the belief in a particular outcome. In this situation, an observation deviating from the prior belief would be less surprising. This property is captured by all the surprise measures presented in this paper.

Another important issue we haven't fully covered is how to determine which surprising events are currently relevant for the particular road user from whose perspective surprise is computed. For example, an unexpected stop on a parallel adjacent road might (if seen) be surprising to the driver, but irrelevant to their current driving task. This ‘‘relevance filtering” speaks to the traditional notion of attention and our road user surprise model needs a similar mechanism for determining what surprising observations are relevant for the current driving task. This involves selecting which other agents should be accounted for in the surprise calculation (e.g., a pedestrian crossing the road unexpectedly in a nearby park is clearly not relevant for a driver on a passing highway), and which actions of those nearby agents are relevant to the subject road user (e.g., lane change away from the subject road user is typically not that relevant even if it is surprising). In other words, we should ignore surprising behavior of nearby road users when the particular behavior has no consequences for our own actions. There are several ways to define such relevance criteria for the surprise model and their detailed evaluation is outside the scope of the present paper. As a general principle, the surprise relevance criteria should select actions for surprise evaluation if they potentially affect the subject road user’s current motion task (e.g., driving, riding, walking) in some way.  

The surprise measures described in this paper have various applications. Here we discuss three main areas: (i) traffic conflict definition, (ii) road user response timing modeling, and (iii) driving behavior evaluation. 

\emph{Traffic conflict definition}: In the traffic conflict literature, measures of spatiotemporal proximity such as time to collision (TTC), required deceleration, post encroachment time (PET) and related metrics are typically used to quantitatively measure traffic conflicts and their severity (\cite{zheng2021modeling, hyden1987development, glauz1980application, ozbay2008derivation}). However, situations with close spatiotemporal proximity are relatively common in everyday driving, while traffic conflicts are non-planned and hence surprising events. For example, an overtaking maneuver in everyday driving with a relatively small TTC to the lead vehicle or a situation where a cyclist intentionally cuts behind a moving car would not generally be considered critical even if the spatiotemporal separation is small. Thus, conditioning traffic conflicts on surprise, such that a conflict needs to involve both spatiotemporal proximity and surprise, potentially reduces the false detection of traffic conflicts. This idea is reflected in the \cite{ISO_2018} definition of a near crash which requires that “The conflict resulting from the trajectory of the conflict partners is not premeditated (planned) by at least one conflict partner”, and is conceptually similar to existing conflict metrics based on “jerky” behavior (\cite{tageldin2016developing, bagdadi2011jerky}). However, combining the type of surprise metrics described here with spatiotemporal proximity is a novel concept not yet explored in the traffic conflict literature. 

\emph{Road user response timing modeling}: Measuring and modeling road user response timing in naturalistic traffic conflicts is challenging, in particular because there is often no clear cut stimulus onset to “start the clock” for a response time measurement. In addition, in normal driving situations, road users often act in anticipation of the stimulus (e.g., slowing down in anticipation of the lead vehicle braking at a red light), in which case the concept of a “response time” makes little sense. \cite{engstrom2022modeling} proposed that to enable a meaningful representation of response timing in naturalistic scenarios, responses to events can be modeled as belief updating in the face of surprising evidence. Based on this idea, a stimulus onset can be defined as the onset of surprising evidence for an event that requires a response (e.g., a traffic conflict), and the response process can be modeled as the gradual accumulation of surprising evidence for the need to respond. Examples of heuristic and computational response timing models based on this idea are given in \cite{engstrom2022modeling} and their application in AV collision avoidance testing is described in \cite{kusano2022collision} and \cite{scanlon2022collision}.         

\emph{Driving behavior evaluation}: Finally, models of surprise can be used more broadly to evaluate the quality of driving behavior, for both human and autonomous drivers. Driving schools teach the importance of driving predictably, and autonomous vehicles should likewise avoid surprising other road users (\cite{de2021driverless}). More broadly, predictable behavior is known as a key factor underlying trust (\cite{lee2004trust}), and hence, the predictability of AV behavior could be expected to underwrite the degree to which they are trusted, both by their direct users and by society as a whole. Our surprise models offer a way to precisely operationalize road user predictability into driving behavior metrics that can be used both offline during AV development and as part of the onboard automated driving system itself\footnote{These, and other techniques may be described in, e.g., U.S. Patent No. 11,447,142; U.S. Patent App. No. 17/946,973; U.S. Patent App. No. 17/399,418; U.S. Patent App. No. 63/397,771; U.S. Patent App. No. 63/433,717; and U.S. Patent App. No. 63/460,815.}.  

We have conceptualized surprise specifically as a violation of expectations of an external state (e.g., another road user’s behavior). However, it should be noted the \emph{active inference} framework suggests a more general notion of surprise based on the idea that (i) the generative model not only predicts external events but also one’s own control actions (e.g., accelerating, steering, braking) and their consequences (e.g., affecting the behavior of other road users) and (ii) the predictions represent the agent’s preferred state (\cite{friston2017active, bruineberg2018free, parr2022active}). From this perspective, surprise can be conceived as any deviation from the predicted (and thus preferred) state of the agent plus environment. For a road user agent, this preferred state may conceptually be characterized as something like “I’m making safe progress towards the destination while respecting rules of the road and other social norms”. According to active inference, the agent’s behavior can then be explained by the single mandate to generate observations that conform to this preferred state, which is equivalent to maximizing the evidence for its generative model or minimizing surprise. Thus, for example, an observed deviation from expected progress towards the destination generates surprise which can be eliminated either by increasing speed (action) or changing the expected progress to align with the observed speed (perception). Our models focus on surprise related to external events, but recent work such as \cite{wei2022modeling, wei2023active, wei2023world} have started exploring this more general notion of surprise which opens up interesting new paths for future road user behavior modeling.

\section{Methods}

We now present the details of Residual Information and Antithesis and compare them to two of the most commonly used surprise measures, surprisal and Bayesian surprise.

\subsection{Residual Information}

Residual Information is a probabilistic mismatch surprise measure which solves a number of practical problems we've encountered when applying common existing  surprise measures to the road traffic domain. One such measure is Shannon information, also known as surprisal (\cite{shannon1948mathematical}).

\begin{equation}
   S(x; P) =  -\log(P(x))
\end{equation}

Many constructs in information theory, including surprisal, assume discrete / categorical probability distributions (\cite{marsh2013introduction}). In our setting however, we are considering a continuous distribution over future position. To apply surprisal, we first need to discretize the distribution:

\begin{equation}
\begin{split}
    P_\varepsilon \triangleq \text{the discretization of $P$ into bins of size $\varepsilon$.}\\
    S_\varepsilon(x; P, \varepsilon) =  S(x; P_\varepsilon) = -\log(P_\varepsilon(x))
\end{split}
\label{eq:discretized_surprisal}
\end{equation}

The first problem with surprisal is that it is non-zero for the most likely outcome. This is inconvenient in practice, and contradicts prior empirical results (\cite{macedo2004modeling}), which found that “the occurrence of the most expected event of the set of mutually exclusive and exhaustive events did not elicit surprise in humans”\footnote{We are explicitly claiming here that surprise and information gain are not equivalent; i.e. there is such a thing as `unsurprising' information.}. The second problem is the choice of the bin size $\varepsilon$; the metric is quite sensitive to it, and diverges to $\infty$ as $\varepsilon$ approaches 0.

\begin{figure}
	\centering
		\includegraphics[width=1\textwidth]{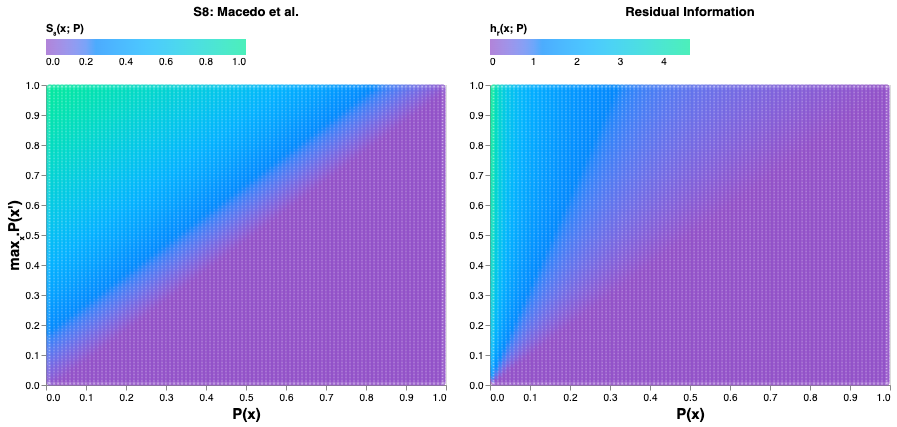}
        \subfloat[\label{fig:residual_info_a}]{\hspace{.45\linewidth}}
        \subfloat[\label{fig:residual_info_b}]{\hspace{.45\linewidth}}
    \caption{$S_8(x; P)$ as a function of $max_{x'}P(x')$ and $P(x)$. The quantities on the diagonal, where $max_{x'}P(x') = P(x)$, are all zero, illustrating the desired zero-floor property. As we approach the top-left of the chart, $P(x)$ becomes lower relative to $max_{x'}P(x')$, leading to a higher surprise value. b) Residual Information, in nats. The isochromatic bands intersecting the origin indicate the scale-invariant nature of the metric.}
	  \label{fig:residual_info}
\end{figure}

\cite{macedo2004modeling} attempted to address the “surprise floor" problem by testing a suite of surprise metrics against the self-reported surprise of study participants presented with {distribution, outcome} pairs. Their most successful metric, $S_8$ (Figure~\ref{fig:residual_info_a}), matched the empirical data better than any of their other formulae, most notably by being 0 when the most likely outcome occurs.          

\begin{equation}
    S_8(x; P) =  \log_2(1+\max_{x'}P(x') - P(x))
    \label{eq:s8}
\end{equation}

\cite{macedo2004modeling} only defined the formula for categorical distributions, so we again discretize $P$ before computing it. As with surprisal, the question of how to select $\varepsilon$ re-emerges; in this case, the metric approaches 0 as $\varepsilon$ approaches 0. Aside from these practical problems, Equation \ref{eq:s8} does not appear to have any information-theoretic interpretation. Table \ref{tab:metric_comparison} summarizes the limitations of these existing metrics.

We created our own probabilistic mismatch surprise measure, Residual Information, to address these shortcomings (Figure~\ref{fig:residual_info_b}).  We first define this measure for a categorical distribution, then demonstrate its generalizability to continuous distributions.

Consider a categorical distribution $P$. We define Residual Information as the difference in information content between the observed outcome and the most likely outcome. This equals zero when the most likely outcome is observed.

\begin{equation}
    h_r(x; P) =  \log(\max_{x'}P(x')) - \log(P(x)) = \log(\max_{x'}P(x')/P(x))
\end{equation}

Now, suppose we wish to apply this formula to a continuous distribution, using the same discretization as for $S_\varepsilon$ and $S_8$. Defining $P_\varepsilon$ as in Equation \ref{eq:discretized_surprisal},

\begin{equation}
    h_r(x; P, \varepsilon) = \log(\max_{x'}P_\varepsilon(x')/P_\varepsilon(x))
\end{equation}

So far, this seems no different than surprisal and $S_8$ in that we are left with the choice of how to set the parameter $\varepsilon$.  However, as $\varepsilon$ approaches 0, $ h_r(x; P, \varepsilon)$ approaches $\log(\max_{x'}P(x')/P(x))$\footnote{ Note that the argument to the log on the right-hand side is a ratio of probability densities from the continuous distribution $P$, rather than masses from the categorical distribution $P_\varepsilon$.}. This means that our formula for $ h_r(x; P)$, which we formulated for the categorical case, generalizes to the continuous case without modification! Table \ref{tab:metric_comparison} summarizes the relative benefits of Residual Information, and its signal quality is highlighted in Figure~\ref{fig:results_surprise}.

\begin{table}[h]
\centering
\begin{adjustbox}{width=1\textwidth}
    \begin{tabular}{|c|c|c|c|}
    \hline
     & Zero-floor & Parameterless & Theoretically meaningful \\
    \hline
    Surprisal &  \textcolor{red}{\xmark}  & \textcolor{red}{\xmark}  & \textcolor{waymogreen}{\cmark} \\
    \hline
     $S_8$ & \textcolor{waymogreen}\cmark & \textcolor{red}{\xmark} & \textcolor{red}{\xmark}\\
    \hline
    Residual Information & \textcolor{waymogreen}\cmark & \textcolor{waymogreen}\cmark & \textcolor{waymogreen}\cmark\\
    \hline
    \end{tabular}
    \end{adjustbox}
    \caption{Probabilistic mismatch surprise metric properties.}
    \label{tab:metric_comparison}
\end{table}

\subsection{Antithesis}

Antithesis is a belief mismatch surprise measure which detects the increased likelihood of a previously unexpected outcome. As described in the previous section, probabilistic mismatch surprise measures detect any observation which was unlikely under our prior beliefs, even if this observation has no bearing on our subsequent beliefs. In contrast, belief mismatch surprise measures specifically detect consequential information with the power to change our beliefs. In our setting, this allows us to measure changes in our predictions about future outcomes, which has the advantage of implicitly considering higher time-derivatives of the predicted quantity. For example, a sudden but significant deceleration will cause a large change in predicted future position, even if it has not yet significantly affected the current position of the vehicle. The same applies to changes in heading or tire angle. This allows us to identify certain surprising actions earlier, as illustrated in Figures~\ref{fig:results_surprise_b} and \ref{fig:results_surprise_d}.

The typical belief mismatch surprise measure found in the literature—Bayesian surprise—is the Kullback-Leibler (KL) divergence between the posterior $P(\bigcdot|y)$ and the prior $P$, (\cite{itti2009bayesian}). In our setting, the predictions are generated at different times, but we take care to compare the predicted distribution over position at a common future time, as illustrated in Figure~\ref{fig:concept_illustration_c}.

\begin{equation}
    D_{KL}(P(\bigcdot|y)||P) = \int P(x|y)\log(P(x|y)/P(x)) dx
\end{equation}

The concern about the zero-floor property, which we discussed in detail in the context of probabilistic mismatch surprise, takes on a different character in the context of belief mismatch measures. Superficially, it appears that KL divergence satisfies the zero-floor property, since when $P = P(\bigcdot|y)$, $D_{KL}(P(\bigcdot|y)||P) = 0$; when applied practically however, this condition is seldom met. The prediction $P(\bigcdot|y)$ is made using additional information $y$ that was not available when prediction $P$ was made. This typically means that uncertainty about the outcome decreases, even if the mode of the prediction does not change. Consequently, $P \neq P(\bigcdot|y)$, and therefore KL divergence is not zero. 

\begin{figure}
	\centering
		\includegraphics[width=1\textwidth]{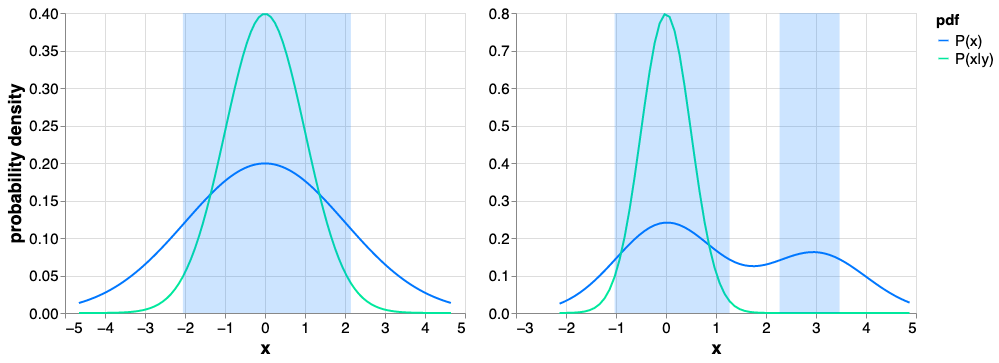}
        \subfloat[\label{fig:antithesis_a}]{\hspace{.45\linewidth}}
        \subfloat[\label{fig:antithesis_b}]{\hspace{.45\linewidth}}
    \caption{Unsurprising information gain.  The blue regions indicate outcomes that are `within expectations' under $P$.  a) Mode-narrowing, corresponding to the acquisition of information confirming a single prior expectation.  b) Mode-removal, corresponding to evidence for one of several plausible but mutually exclusive outcomes.}
	  \label{fig:antithesis}
\end{figure}

Moreover, consider a vehicle driving down the highway with its turn signal on. Are they about to change lanes? Did the driver forget to turn off their signal? Both outcomes are within expectations, therefore evidence for either of these hypotheses is unsurprising. Figure~\ref{fig:antithesis_b} illustrates this scenario in caricature. On the other hand, suddenly slamming the brakes to avoid a previously unseen pedestrian may surprise the driver of the following vehicle quite profoundly, as illustrated in Figures~\ref{fig:scenarios_pics_c} and \ref{fig:scenarios_pics_d}, and Figures~\ref{fig:results_surprise_c} and \ref{fig:results_surprise_d}.

We designed Antithesis to silence “unsurprising" information gain such as mode-removal and mode-narrowing (Figure~\ref{fig:antithesis}).

\begin{equation}
\begin{split}
    C(P, x, y) = \log(P(x)) < E_{x'}[\log(P(x'))] \wedge P(x|y) > P(x)\\
    A(y; P) = \int_{C(P, x, y)} P(x|y)\log(P(x|y)/P(x))dx
\end{split}
\label{eq:antithesis}
\end{equation}

Equation \ref{eq:antithesis} means that we evaluate the KL integral only over the region where the predicate $C$ is true. When using sampling methods to compute the integral, this corresponds to evaluating $C$ for each sample, and discarding all the samples for which it is false.

C is composed of two conditions: (i) the “outside expectations" condition $\log(P(x)) < E_{x'}[\log(P(x'))]$ and (ii) the “increased belief" condition $P(x|y) > P(x)$. Together, these conditions restrict the domain of the integral to regions representing an increased likelihood of a previously unexpected outcome: an “antithesis" which opposes the original hypothesis.

Our definition of “within expectations" is—loosely speaking—that the information content of the observation is below average for the distribution.  Alternative definitions are certainly possible; one can imagine parameterizing the metric on this threshold to tune its sensitivity, for example. Empirically, we find that Antithesis is zero more often than KL divergence, due to the “outside expectations" condition.  Consequently, Antithesis has more power to distinguish between surprising and unsurprising events.

\section{Declaration}
The ideas discussed in this manuscript may be described in patents filed by Waymo, e.g., U.S. Patent No. 11,447,142; U.S. Patent App. No. 17/946,973; U.S. Patent App. No. 17/399,418; U.S. Patent App. No. 63/397,771; U.S. Patent App. No. 63/433,717; and U.S. Patent App. No. 63/460,815.

%% Loading bibliography style file
%\bibliographystyle{model1-num-names}
\bibliographystyle{cas-model2-names}

% Bibliography
%\newpage
% \bibliographystyle{numeric}
\bibliography{references}

\end{document}